%% file: samplepaper.tex
\documentclass[runningheads]{llncs}
\usepackage[T1]{fontenc}
\usepackage{graphicx}

\usepackage{times}
\usepackage{latexsym}
\usepackage{enumerate}

\usepackage[utf8]{inputenc}
\usepackage{tikz}
\usepackage{caption}
\usepackage{subcaption}

\usepackage{microtype}
\usepackage{diagbox}
\usepackage{inconsolata}
\usepackage{multirow}
\usepackage{subcaption}
\usepackage{arydshln}
\usepackage{booktabs}

\usepackage[flushleft]{threeparttable}
\usepackage{scalerel,xparse}
\usepackage{hyperref}
\usepackage{color}
\usepackage{wrapfig}

\begin{document}
\title{Overview of the NLPCC 2025 Shared Task: Gender Bias Mitigation Challenge}
\author{
Yizhi Li\inst{1,2} \and
Ge Zhang\inst{1} \and
Hanhua Hong\inst{2} \and
Yiwen Wang\inst{3} \and 
Chenghua Lin\inst{2}
}
\authorrunning{Y. Li et al.}
\institute{
Multimodal Art Projection Research Community \and
University of Manchester  \and
Independent Researcher 
}
\maketitle              %
\begin{abstract}
As natural language processing for gender bias becomes a significant interdisciplinary topic, the prevalent data-driven techniques, such as pre-trained language models, suffer from biased corpus. 
This case becomes more obvious regarding those languages with less fairness-related computational linguistic resources, such as Chinese. 
To this end, we propose a Chinese cOrpus foR Gender bIas Probing and Mitigation (\textbf{CORGI-PM}), which contains \textbf{32.9k} sentences with high-quality labels derived by following an annotation scheme specifically developed for gender bias in the Chinese context. 
It is worth noting that CORGI-PM contains 5.2k gender-biased sentences along with the corresponding bias-eliminated versions rewritten by human annotators.
We pose three challenges as a shared task to automate the mitigation of textual gender bias, which requires the models to detect, classify, and mitigate textual gender bias. 
In the literature, we present the results and analysis for the teams participating this shared task in NLPCC 2025.

\end{abstract}
\section{Introduction}
There is a growing consensus that the identification and prevention of toxic gender attitudes and stereotypes are of vital importance for society \cite{blodgett2020language}. Given that gender-biased information can be presented and disseminated extensively in textual form, it is of paramount significance to develop automatic approaches for detecting and alleviating textual gender bias.
As the blooming of open-source models and training corpus \cite{gao2020pile,bai2023qwen,zhang2024map,wake2024yi,young2024yi}, Natural language processing (NLP) community has found extensive applications in text-related scenarios and has exerted a considerable influence on gender bias issues~\cite{costa2019analysis}. On one hand, pre-trained language models (LMs), which serve as a crucial technique in modern NLP, have been demonstrated to absorb the subjective gender bias present in the internet-wise training corpus and even have the potential to amplify it \cite{zhao-etal-2017-men,kotek2023gender,wan2023kelly,dong2024disclosure}. On the other hand, the application of cutting-edge NLP techniques for exploring and mitigating gender bias holds increasing promise. 
This status in quo necessitates the construction of high-quality corpora to facilitate the research for gender bias mitigation.

\input{fig/fitler_pipeline}

Building a high-quality text corpus has been one of the key tangents in improving NLP applications for debiasing gender stereotypes in texts~\cite{sun2019mitigating}. Some researchers introduce \textit{automatic} annotation techniques, such as gender-swapped based methods, to create corpora for gender bias mitigation~\cite{lu2020gender,zhao2018learning,rudinger2018gender}. While it is attractive to build a large-scale corpus without heavy labor, automatic gender-swapped based methods highly depend on the quality of base language models and are prone to creating nonsensical sentences~\cite{sun2019mitigating}. 
To address this issue, some works devote effort to developing \textit{human-annotated} corpora for gender bias mitigation. 
However, these corpora either mainly focus on word- or grammar-level bias~\cite{webster2018mind,zhu2020great,sahai2021predicting,zhou2019examining}, without considering the nuanced biases appearing in specific contexts. Or some of the corpora concentrate only on sexism-related topics~\cite{parikh2019multi,chiril2020annotated,chiril2021nice,jiang2022swsr}. 
Moreover, existing works on gender bias exclusively focus on English~\cite{costa2019analysis}, where few datasets exist for other influential languages such as Chinese. We aim to tackle the aforementioned issues by providing a high-quality Chinese corpus for contextual-level gender bias probing and mitigation with a hybrid pipeline including automatic corpus filtering and human annotation. 

To this end, we propose the Chinese cOrpus foR Gender bIas Probing and Mitigation (\textbf{CORGI-PM}) dataset, which consists of \textbf{32.9k} human-annotated sentences, including both gender-biased and non-biased samples. 
To construct the initial candidate pool for further processing, we propose an automatic method to build a potentially gender-biased sentence set from existing large-scale Chinese corpora, as illustrated in Fig.~\ref{retriving_pipeline}. 
Inspired by the metric leveraging language models for gender bias score calculation proposed in~\cite{Bolukbasi2016ManIT}, the samples containing words of high gender bias scores are first recalled at this stage.
Following the retrieval, the samples are then reranked and filtered according to their sentence-level gender-biased probability~\cite{jiao2021gender}. 
After the candidate pool construction, we select human annotators with qualified educational backgrounds and design an annotation scheme to label the acquired sentences. 

During the annotation, the candidates are classified into biased and non-biased classes, and the biased samples are further categorized into three fine-grained categories considering their contexts.
To facilitate research in bias mitigation research, we also require the annotators to paraphrase the biased sentences into gender-neutral ones while maintaining invariant semantics to build a parallel subset corpus.
Based on the labeled corpus, we further pose three challenges in CORGI-PM, \textit{i.e.}, gender bias \textbf{detection}, \textbf{classification}, and \textbf{mitigation}, which come with clear definitions and evaluation protocols for NLP tasks in gender bias probing and mitigation. 
In order to provide referential baselines and benchmarks for our proposed challenges, we conduct random data splitting with balanced labels and implement experiments on pre-trained language models in zero-shot, in-context learning, and fine-tuning paradigms. We discuss the experimental settings and provide result analysis in \S\ref{sec:chanllenge}.

In summary, we provide a well-annotated Chinese corpus for gender bias probing and mitigation, along with clearly defined corresponding challenges. With a properly designed annotation scheme, CORGI-PM provides a corpus of high quality that assists models in detecting gender bias in texts. More importantly, other than the 22.5k human-annotated non-biased samples, all the 5.2k biased sentences in our corpus are further labeled with gender bias subclasses and companies with parallel bias-free versions provided by the annotators\footnote{To eliminate the data contamination, we additionally provide 100 samples for each task in addition to the original splits~\cite{zhang2023corgipmchinesecorpusgender}.}. 

\section{Related Work}

\paragraph{Gender Bias Corpus.} \quad High-quality gender bias corpus, especially contextual-level gender bias corpus, is significant for mitigating gender bias contained in language models, but hard to collect. 
Previous research work widely uses gender-swapping to build a corpus for evaluation or mitigate gender bias in LMs \cite{zhao-etal-2018-gender,kiritchenko-mohammad-2018-examining}.  
Metrics and corpus are introduced to measure gender bias in abusive languages as well \cite{jiang2022swsr,parikh-etal-2019-multi,Dixon-2019-etal-MitigatingBias}. 
Word-level Chinese gender bias corpora have been proposed about adjectives \cite{zhu-liu-2020-wei}, and careers \cite{srivastava2022beyond}. 
\cite{zhao-etal-2018-gender} also shares an automatically constructed Chinese sentence-level gender-unbiased data set.
In sharp contrast, CORGI-PM is the first human-annotated Chinese gender bias detection dataset, one of the rare gender bias classification datasets \cite{parikh-etal-2019-multi}, and the first human-annotated gender-bias correction dataset. 

\paragraph{Bias Detection and Correction.} \quad Different Gender Bias Evaluation Testsets(GBETs) have been designed to automatically evaluate models trained for specific tasks for gender bias \cite{sun-etal-2019-mitigating,kiritchenko-mohammad-2018-examining,webster-etal-2018-mind}. 
Word-swapping is the most widely used technique for gender bias correction and widely used to mitigate the gender bias in hate-speech \cite{park-etal-2018-reducing}, regional bias mitigation\cite{li-etal-2022-herb}, knowledge graph \cite{pmlr-v81-madaan18a}, sentiment analysis \cite{kiritchenko-mohammad-2018-examining}, and the general language model \cite{zhao-etal-2017-men}. As a comparison, CORGI-PM serves as a manual standard dataset for testing LMs' ability to measure and mitigate gender bias.

\section{Data Collection}
\subsection{Sample Filtering}
We propose an automatic processing method to recall, rerank, and filter annotation candidates from raw corpora using a two-stage filtering from word-level to sentence-level, as illustrated in Fig.~\ref{retriving_pipeline}. The Chinese sentence samples are mainly screened out from the SlguSet~\cite{zhao2021} and the CCL corpus~\cite{zhan2019CCL}.
To recall gender-biased words or retrieve candidate sentences with gender bias scores, we compare the target word/sentence representations with the \textit{seed direction}, which can be calculated by the subtraction between the word embeddings of \texttt{she} and \texttt{he} ~\cite{Bolukbasi2016ManIT,jiao2021gender}.
We leverage different Chinese LMs including ERNIE \cite{zhang2019ernie}, CBert \cite{cui2020revisiting}, and Chinese word vectors \cite{qiu2018revisiting} to acquire the word-level and sentence-level representations. For word-level filtering, we use the mentioned metric to build a vocabulary of high bias scores and recall sentences containing such words from the raw corpora with exact matches. We compute gender bias scores of the crawled sentences and group them by the gender bias keywords acquired in the previous stage for sentence-level filtering. The final sentences for annotation are then selected according to a specific global threshold score and an in-group threshold rank. 

\begin{table}[bt]

\centering
\scalebox{1.0}{
    \begin{tabular}{l:c|ccc}
    \toprule
\multicolumn{2}{c|}{\textbf{Sample}} &  \multicolumn{3}{c}{\textbf{Quantity}} \\ 
    \cmidrule{3-5}
    \multicolumn{2}{c|}{\textbf{Category}} & Train  & Valid  & Test \\
    \midrule \midrule

    \parbox[t]{2mm}{ \multirow{3}{*}{\rotatebox[origin=c]{90}{\textbf{Biased}}}}& AC  & 1.90k & 235 & 237 \\
     
       & DI &  2.70k & 334 & 337 \\
       & ANB &  2.47k & 306 & 309 \\

    \midrule
    \multicolumn{2}{c|}{\textbf{Non-biased}} & 21.4k & 516  & 526 \\
    \midrule

    \multicolumn{2}{c|}{\textbf{Overall}} & 30.1k & 1391 & 1409 \\
  
    \bottomrule
    \end{tabular}
}
\caption{
Overall Statistics of the CORGI-PM Dataset. 
The notations, \textbf{AC}, \textbf{DI}, and \textbf{ANB} represent specific bias labels described in \S~\ref{sec:annotaion_schem}.
}
\vspace{-10mm}
\label{overall_dataset_stat}
\end{table}

\begin{table*}[!ht]

\centering
\scalebox{1.0}{
    \begin{tabular}{c|ccc|ccc|ccc}
    \toprule
    \multicolumn{1}{c|}{\textbf{Linguistic}} &   \multicolumn{3}{c|}{\textbf{Non-biased}} & \multicolumn{3}{c|}{\textbf{Biased}} & \multicolumn{3}{c}{\textbf{Corrected Biased}} \\ 
    \midrule
    \multicolumn{1}{c|}{\textbf{Info.}} &  Train  & Valid  & Test & Train  & Valid  & Test & Train  & Valid  & Test \\ 
    \midrule \midrule
    Word & 724k  & 18.9k & 17.7k & 228k & 24.8k & 28.3k & 265k & 27.1k & 30.0k\\  
    Dictionary & 574k  & 14.4k & 14.1k & 167k & 18.4k & 20.4k & 191k & 19.9k & 21.5k\\  
    Character &  1,156k & 30.1k & 28.1k & 358k & 39.2k & 44.4k & 417k & 42.8k & 46.9k\\
    Sent. Length & 53.952 & 58.397 & 53.473 & 85.837 & 76.087 & 85.214 & 99.839 & 82.853 & 89.939 \\
    \bottomrule
    \end{tabular}
}
\caption{Linguistic Characteristics of the Corpus.
\textit{Word, Dictionary, and Character} separately denote the total Chinese word number, total unique Chinese word number, and total character number of the specific categories. The sentence lengths are defined as the number of containing characters.}
\label{corrected_stat}
\vspace{-10mm}
\end{table*}

\subsection{Annotation Scheme}\label{sec:annotaion_schem}

The annotation scheme is designed for gender bias probing and mitigation. For gender bias probing, the annotators are required to provide the following information given a sentence: whether gender bias exists; if so, how the bias is established. For gender bias mitigation, the corrected non-biased version of the biased sentences is also required. We further describe the annotation scheme details in the following paragraphs.

\paragraph{Existence and Categorization.}
The annotators are required to annotate whether the sentence is gender-biased (\textbf{B}) or non-biased (\textbf{N}) in contextual-level or word-level, and further clarify how the bias is established.
Given that our raw data is collected using gender-related keywords or from gender-related corpus, the samples annotated without gender bias are useful human-annotated negative samples for detecting gender bias. 
To additionally provide information about gender bias categorization, we classify gender bias types into three subtypes : (1) Gender Stereotyped activity and career choices \textbf{(AC)}; (2) Gender Stereotyped descriptions and inductions \textbf{(DI)}; and (3) Expressed gender-stereotyped attitudes, norms and beliefs \textbf{(ANB)}. The classification standard is inspired by \cite{king2021gender} and further summed up into the mentioned subtypes.  

\begin{figure}[!ht]
    \centering
    \includegraphics[width=0.95\linewidth]{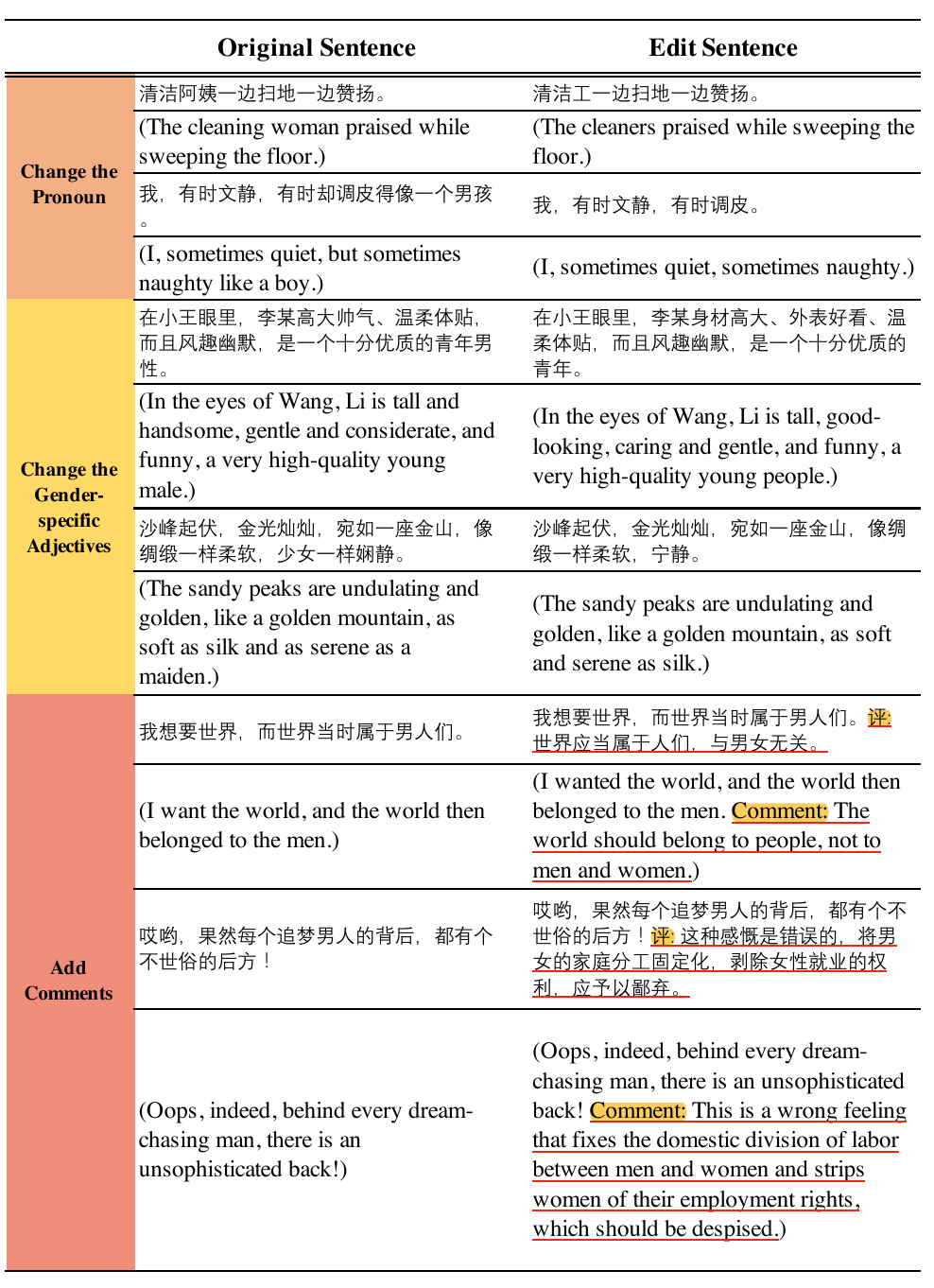}
    \caption{Case Study of Mitigation Annotation Patterns.}
    \label{fig:CaseStudyEditing}
\end{figure}

\paragraph{Bias Mitigation.}
Annotators are further required to mitigate the gender bias of selected sentences while keeping the original semantics. 
We also ask our annotators to diversify the expressions if applicable.
The major revision patterns can be summarized as follows: 
(1). \textit{Replace} the gender-specific pronouns with neutral pronouns. 
(2). \textit{Replace} the gender-specific adjectives with neutral descriptions with similar semantics definitions. 
(3). \textit{Add} additional comments to neutralize the sentences that cannot be directly mitigated.

As shown in Fig.~\ref{fig:CaseStudyEditing}, the three mitigation revision patterns could cover various cases in the selected sentences.
These three major mitigation annotation patterns are not used exclusively in the annotation but can be used in combination.
Except for the three mentioned patterns, we require the annotators to apply several other linguistic skills to cover corner cases, including deleting gender-specific pronouns and replacing vehicles in gender-related metaphors.

\section{CORGI-PM Corpus Analysis}
\label{sec:corpusAnalysis}

In this section, we report the linguistic statistics of CORGI-PM as Tab.~\ref{overall_dataset_stat}. %
We conduct a balanced splitting to create the valid and test set considering the negative-positive ratio and bias subclass proportion in the global distribution. 
As revealed in Tab.~\ref{corrected_stat}\footnote{We use the {\texttt{Jieba} (\url{https://github.com/fxsjy/jieba})} to parse.}, we observe two major differences between the biased sentences and the corresponding human debiased ones: the debiased are usually longer and have more diverse expressions, revealed by the sentence length and the vocabulary size metrics. 
Except for the samples debiased with additional neutralization comments, we hypothesize that it is due to human annotators' intention to keep the semantic information unchanged and the sentence coherent while mitigating gender bias. In this case, they may use more conjunctions and longer descriptions compared to some gender-biased inherent expressions.

\paragraph{Quality Monitoring and Control.}\label{sec:appen_qm}
We used a standardized operating method and educated our annotators to achieve high-quality annotations as follows:

(1). \textbf{Annotators} \quad We have 6 annotators, which were all native speakers of Chinese.
Annotators were only qualified to do the annotation if they went through several societal \cite{king2021gender,xu2019cinderella} and computer science research works \cite{sun2019mitigating,zhao2018learning} about gender bias before the annotation procedure. 
All annotators held a bachelor's degree. 
\cite{waseem2016racist} points out that expert annotators are more cautious and can improve the corpus quality with a large margin, which proves the necessity of our training procedure. 
We also kept the number of male and female annotators equal.

(2). \textbf{Gender Equality of Raw Corpus} \quad In the raw data collection procedure, we keep the number of man-related keywords and woman-related keywords equal and make the number of samples recalled according to different keywords balanced. As a result, the raw data and the final data should hold gender equality.

(3). \textbf{Annotation Procedure} \quad Our annotation procedure is separated into two stages. 
In the first stage, annotators are encouraged to not enter any samples that they are not certain about. 
In the second stage, we have annotators cross-checking annotations.
We did not enter any contradictory samples.

(4). \textbf{Inter-annotator Agreement} \quad Given the domain and purpose of the dataset, we want to build the dataset as high quality as possible. After an initial annotation round with 6 annotators, we also report inter-annotator agreement in Tab.~\ref{iaa}. to verify annotation reliability, where the IAA among three annotators on bias classification, detection, and mitigation is 0.802, 0.935, and 0.987, respectively.

\begin{table}[h!]

\centering
   \scalebox{1.0}{
    \begin{tabular}{cccc}
    \toprule
     & \textbf{Classification} & \textbf{Detection} & \textbf{Mitigation}\\
    \midrule\midrule
    \multicolumn{1}{c}{\textbf{IAA}} & 0.802& 0.935 & 0.987\\
    \bottomrule
    \end{tabular}
    }
    \caption{Inter-Annotator Agreement (IAA)}
    \label{iaa}
\vspace{-10mm}
\end{table}

\paragraph{Word Cloud Analysis.}
\label{sec:appen_cor_cloud}
We provide word cloud analysis of Ernie and Chinese-Electra in the section about adjectives and career words.
More available word cloud analysis will be available in our public repository.
The words are ranked according to the absolute value of their gender bias score calculated along the method used by \cite{Bolukbasi2016ManIT,jiao2021gender}. 
There is a noticeable word-level gender stereotype according to the word cloud. 
For example, a man is robust and a woman is motherly, a man is suitable for a fitness instructor and a woman is suitable for a choreographer.
We also conduct word cloud analysis for language models pre-trained by different corpora.

\section{Derived Gender Bias Mitigation Challenges}\label{sec:chanllenge}

To provide a clear definition for automatic textual gender bias probing and mitigation tasks, we propose corresponding challenges and standardize the evaluation protocols. 
We address two tasks, bias detection and classification, for gender bias probing to evaluate how well the language models can distinguish the biased contexts.
Furthermore, we formalize the gender mitigation challenge as a sentence correction task to benchmark the bias mitigation ability of the language models.

\subsection{Challenges of Detection and Classification}\label{sec:chanllenge_DC}

\paragraph{Definition.} 
We regard both the gender bias detection and classification challenges as \textit{supervised classification} tasks and evaluate them with metrics of consensus.
The gender bias detection challenge can be regarded as a binary classification task, where the model is required to predict the probability that a given sentence contains gender bias. As described in \S~\ref{sec:annotaion_schem}, biased samples are further categorized into one or more kinds. Therefore, we can address the gender classification challenge as a multi-label classification task. The precision, recall, and F1-score are selected as the main metrics in these two challenges. Class-wise metrics and macro average summarized evaluation are required through both valid and test sets to show the performance of language models.

\paragraph{Experiment Settings.}
Then we use the test sets to perform a classification query on the saved file. The processing time for the classification of gender bias is approximately 1 hour. We calculated the precision, recall, and F1 score to analyze model performance. 
We test the performance on both "yes" and "no" detection.

\subsection{Challenge of Mitigation}

\begin{wraptable}{r}{0.5\textwidth}
    \begin{tabular}{ccc}
    \toprule
    \textbf{Task} & \textbf{Team} & \textbf{Performance}\\
    \midrule
    \multirow{6}{*}{Detection} & ZZU-NLP & .850 \\
    & ZZU-NLP\_DS & .741 \\
    & Cloud Lab & .720 \\
    & YNU-HPCC & .714 \\
    & Prompt & .712 \\
    & \underline{MAP-Neo} & .653 \\
    \midrule
    \multirow{6}{*}{Classification} & ZZU-NLP & .646 \\
    & Team0071 & .548 \\
    & Hu & .531 \\
    & \underline{MAP-Neo}& .526 \\
    & YNU-HPCC & .509 \\
    & Prompt & .505 \\
    \midrule
    \multirow{6}{*}{Mitigation} & ZZU-NLP & .294 \\
    & Team0071 & .293 \\
    & YNU-HPCC & .293 \\
    & Prompt & .288 \\
    & IR901 & .271 \\
    & \underline{MAP-Neo} & .091 \\
    \midrule
    \multirow{6}{*}{Overall} & ZZU-NLP & .597 \\
    & YNU-HPCC & .505 \\
    & Prompt & .502 \\
    & Cloud Lab & .479 \\
    & ZZU-NLP\_DS & .472 \\
    & \underline{MAP-Neo} & .423 \\
    \bottomrule
    \end{tabular}
    \caption{The leaderboards for all tasks. Our baseline model is underscored.}
    \label{tab:leaderboard}
    \vspace{-20mm}
\end{wraptable}
\paragraph{Definition.} The gender bias mitigation challenge can be regarded as a sentence correction task, where the model is required to generate a non-biased version of a biased sentence.
As a natural language generation task, the model outputs could be evaluated by N-gram based metrics.

\paragraph{Experiment Settings.}
Regarding the decision on \textit{evaluation metrics}, we conduct extensive human evaluations on the debiased sentences considering both gender bias and coherence aspects. 
In human evaluation, we shuffle the debiased written by humans and different models, and asked annotators to grade the results using the answer range from 1-\textit{not at all} to 7-\textit{extremely gender biased/extremely fluent} without providing the information of the source.
Additionally, we use automated reference-based metrics to evaluate the bias mitigated sentences~\cite{sharma2017nlgeval}, including BLEU \cite{papineni2002bleu}, ROUGE-L \cite{lin2004rouge}, and METEOR \cite{agarwal2007meteor}. 

\section{Leaderboard}

Table \ref{tab:leaderboard} presents the leaderboards for all the tasks, listing only the top-performing teams and our baseline model MAP-Neo. For each task, a representative evaluation metric is selected to compare model performance across teams. Specifically, for the detection and classification tasks—both formulated as binary classification problems—the F1 score is used as the primary performance metric. For the mitigation task, model performance is assessed using the average of BLEU, METEOR, and ROUGE-L F1 scores. An overall ranking is then derived by averaging each team's scores across the three tasks. The results demonstrate that Team ZZU-NLP achieved the highest performance in all three tasks, thereby securing the top overall position in the shared task. 

Notably, while many teams achieved strong results in the detection task, performance declined considerably in the classification task, with some models even falling below the baseline. Furthermore, the performance remain consistently low across all submissions for the mitigation task. This performance discrepancy supports the validity of our stepwise three-task framework, which introduces increasing complexity in a reasonable and structured difficulty progression. Furthermore, the outcomes underscore the persistent challenges in accurately identifying and effectively mitigating gender bias in Chinese-language texts. They highlight the need for models to possess both a deep understanding of gender bias and advanced linguistic capabilities to identify and appropriately revise gender-biased content.

\section{Conclusion}

We introduce CORGI-PM, the first Chinese human-annotated corpus designed specifically for probing and mitigating gender bias in text. This corpus also serves as the basis for defining and evaluating metrics across three challenges, aimed at testing the performance of state-of-the-art language models in detecting, classifying, and mitigating textual gender bias. Our proposed challenges are intended to establish benchmarks for assessing the capabilities of language models in this domain.
Our experiments and the submissions demonstrate that the fine-grained subclass labels in our sentences enhance the models' ability to probe for gender bias, while our parallel, human-written debiased data provides robust supervision for training generative language models. We also analyze the performance of submissions from participating teams. The results highlight the persistent challenges in identifying and mitigating gender bias in Chinese corpora, underscoring the value of our work.
In conclusion, we suggest that future research utilizing CORGI-PM could significantly advance the field of NLP by improving methods for both probing and mitigating gender bias in textual data.

\section*{Limitations}
There are several major limitations in this research work. 
Due to the high requirement of annotators for annotating gender-biased sentences and correcting such sentences, we only choose annotators with higher education, which may lead to potential cognitive bias. 
In addition, CORGI-PM mainly focuses on gender bias but has not explored the bias phenomenon across different domains or disciplines, which could be potentially derived from the comprehensive benchmark for LLMs \cite{wang2024mmlu,du2025supergpqa} .

\bibliographystyle{splncs04}
\bibliography{aaai25}
\end{document}

%% file: fig/fitler_pipeline.tex
\begin{figure}[!hbt]
\centering

\includegraphics[width=0.5\linewidth]{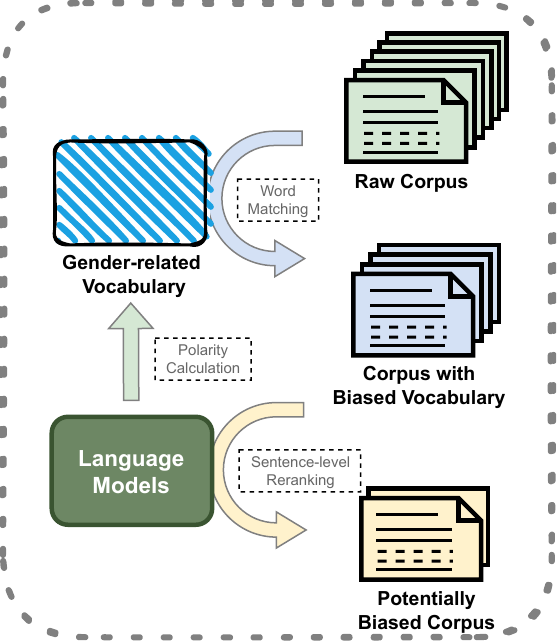} 
\caption{Pipeline of Retrieving and Filtering Potentially Biased Sentences Candidate Pool from Raw Corpus for Human Annotation. }
\vspace{-7mm}
\label{retriving_pipeline}
\end{figure}